\pdfoutput=1

\documentclass[11pt]{article}

\usepackage{acl} 

\usepackage{times}
\usepackage{latexsym}

\usepackage[T1]{fontenc}

\usepackage[utf8]{inputenc}

\usepackage{microtype}

\usepackage{multirow}
\usepackage{tabularx}
\usepackage{graphicx}
\usepackage{booktabs}
\usepackage{paralist}

\usepackage[table]{xcolor}

\usepackage{listings}
\usepackage{paralist}

\colorlet{punct}{red!60!black}
\definecolor{background}{HTML}{EEEEEE}
\definecolor{delim}{RGB}{20,105,176}
\colorlet{numb}{magenta!60!black}

\lstdefinelanguage{json}{
    basicstyle=\normalfont\ttfamily,
    numbers=left,
    numberstyle=\scriptsize,
    stepnumber=1,
    numbersep=8pt,
    showstringspaces=false,
    breaklines=true,
    frame=lines,
    backgroundcolor=\color{background},
    literate=
     *{0}{{{\color{numb}0}}}{1}
      {1}{{{\color{numb}1}}}{1}
      {2}{{{\color{numb}2}}}{1}
      {3}{{{\color{numb}3}}}{1}
      {4}{{{\color{numb}4}}}{1}
      {5}{{{\color{numb}5}}}{1}
      {6}{{{\color{numb}6}}}{1}
      {7}{{{\color{numb}7}}}{1}
      {8}{{{\color{numb}8}}}{1}
      {9}{{{\color{numb}9}}}{1}
      {:}{{{\color{punct}{:}}}}{1}
      {,}{{{\color{punct}{,}}}}{1}
      {\{}{{{\color{delim}{\{}}}}{1}
      {\}}{{{\color{delim}{\}}}}}{1}
      {[}{{{\color{delim}{[}}}}{1}
      {]}{{{\color{delim}{]}}}}{1},
}

\title{A Multilingual Corpus of Socio-political News for Event Coreference Resolution}
\title{A Multilingual Corpus for Socio-political Event Coreference Resolution}
\title{A Multilingual Corpus for Event Coreference Resolution \\ for Social Sciences}
\title{Event Coreference Resolution for Contentious Politics Events}

\author{Ali Hürriyetoğlu\textsuperscript{$\star$}, 
Osman Mutlu\textsuperscript{$\star$}, 
Fatih Beyhan\textsuperscript{$\diamond$}, \\
\textbf{Fırat Duruşan\textsuperscript{$\star$},
Ali Safaya\textsuperscript{$\star$},
Reyyan Yeniterzi\textsuperscript{$\diamond$},
Erdem Y{\"o}r{\"u}k\textsuperscript{$\star$}}
 \\
  \textsuperscript{$\star$}Ko\c{c} University, \textsuperscript{$\diamond$}Sabancı University \\
 {\tt \{ahurriyetoglu,omutlu,fdurusan,asafaya19,eryoruk\}@ku.edu.tr} \\
 {\tt \{fatihbeyhan,reyyan.yeniterzi\}@sabanciuniv.edu}
}

\begin{document}
\maketitle
\begin{abstract}
We propose a dataset for event coreference resolution, which is based on random samples drawn from multiple sources, languages, and countries.
Early scholarship on event information collection has not quantified the contribution of event coreference resolution.
We prepared and analyzed a representative multilingual corpus and measured the performance and contribution of the state-of-the-art event coreference resolution approaches.
We found that almost half of the event mentions in documents co-occur with other event mentions and this makes it inevitable to obtain erroneous or partial event information. We showed that event coreference resolution could help improving this situation.
Our contribution sheds light on a challenge that has been overlooked or hard to study to date. Future event information collection studies can be designed based on the results we present in this report. The repository for this study is on \url{https://github.com/emerging-welfare/ECR4-Contentious-Politics}.
\end{abstract}


\section{Introduction}

Event databases are of great utility in research projects in various fields of social sciences. Social actions of groups and individuals, contentious or cooperative interactions between states and societies, and among various social groups all manifest themselves as events. Thus, event data are crucial in understanding a wide variety of social and political phenomena such as modes of political participation, patterns of migration, and social and political conflict. As any type of data that serves as a source of scientific variables, completeness and reliability of event data have direct bearing on the rigor of these studies. Indeed, since many sociological, political scientific, or economic analyses that rely on event databases also inform policy, it is arguable that quality of research has indirect bearing on the well-being of citizens in some manner. This makes maximizing the quality of event databases even a worthier goal. 


Social scientists have long been working on creating automated event databases. Conflict and Mediation Event Observations (CAMEO)~\cite{Gerner+02}, Integrated Data for Events Analysis (IDEA)~\cite{Bond+03}, and PLOVER\footnote{\url{https://github.com/openeventdata/PLOVER}, accessed on October 10, 2021.} have been the main proposals of event characterizations in social sciences. Semi-automatic~\cite{Nardulli+15} and automated approaches~\cite{Leetaru+13,Boschee+13,Schrodt+14,Sonmez+16,Hurriyetoglu+19b,Hurriyetoglu+19a} have been developed by adopting these formalisms. 

At the same time, the NLP community has achieved some consensus on the treatment of events both in terms of task definition and appropriate techniques for their detection~\cite{Pustejovsky+05,Doddington+2004,Song+15,Getman+18}. However, in order to be useful for social scientists, these formalisms, related language resources, and the automated systems that realize them need to be adjusted or extended in relation to certain cases. For instance the details of the event descriptions and sampling of the documents in the datasets that demonstrate application of these formalisms should reflect the richness and nuances of the events as they are reported in various social and political contexts, dialects, and languages. Moreover, the sampling of the documents to be annotated plays a critical role in determining and prioritizing linguistic characteristics that the automated approaches should handle.

The results yielded by approaches of both communities to date are either not of sufficient quality, require tremendous effort to be replicated with both in- and out-of- distribution data, are immeasurable in terms of quality as there is not any gold standard list of events, or is not comparable to each other ~\cite{Wang+16,Ward+13,Ettinger+17,Plank16,Demarest+2018}. 




Any new project for creating an event database in this line still finds itself making design decisions such as using only the heading sentences in a news article~\cite{Johnson+2016} or not considering event coreference information~\cite{Boschee+13,Tanev+08} without being able to quantify the effect of these decisions on quality of the output. \citet{Weischedel+18} assume that event coreference information may not be necessary for forecast model creation because the number of mentions in the news may already be a useful surrogate for some forecasting models. However, the same opinion piece was concluded by acknowledging the value of the event-event relation information. Therefore the effect of incorporating event-event information on use cases in social sciences domain still remains an open issue.

The event coreference, which is in-document in the scope of our study, identification is the least studied phenomenon by both NLP and social scientists. There are still many unknowns, which are either overlooked or ignored, about this phenomenon~\cite{lu-ng-2021-conundrums}. More information in this respect will enable the creation of precise and complete event databases by decreasing the amount of duplication and partiality of event information~\cite{zavarella2020mastering}. The following are only the first set of questions that should be responded in order to proceed in quantifying event coreference and improve our methodology for event information collection. What is the number of events in a news report in average? How is the information about an event is spread in a document? How information about multiple events co-occurs in a report? What is the prevalence of the expressions that refer to multiple events? How frequently sentences contain information related to multiple events? Does occurrence of event coreference differ across languages? What is the ratio of the documents and events that can benefit from event coreference resolution in a random sample? How do state-of-the-art text processing tools perform on the event coreference task? This report provides answers to majority of these questions by providing a new event coreference corpus that is created by exploiting news articles drawn from various contexts randomly and using a recall-optimized active learning approach. We also demonstrate the performance of various baseline and state-of-the-art approaches to tackle the event coreference resolution task utilizing this corpus.



We present related work in Section~\ref{sec:relwork}. Next, the protest event definition, the methodology we applied to create the corpus, and the corpus characteristics are provided in the Sections~\ref{sec:protest-definition}, ~\ref{sec:methodology-corpus-create}, and~\ref{sec:corpus-character}. The Section~\ref{sec:event-sep} describes the conditions that lead us to consider events as the same or separate events. Our effort for tackling event coreference resolution and the results are demonstrated in the Sections~\ref{sec:event-coref-resolution} and~\ref{sec:results}. Finally, the Section ~\ref{sec:conclusion} concludes this report. 




\section{Related work}
\label{sec:relwork}

The event coreference resolution task was first introduced in the scope of MUC 6~\cite{grishman-sundheim-1996-message} and MUC 7~\cite{chinchor-1998-overview} as a template filling task. Although it was not an explicitly specified task, identifying whether events are coreferent or not was a key component in this task, as it directly affects the number of templates to be filled. Automatic Content Extraction (ACE 2005) dataset~\cite{Doddington+2004}, ECB~\cite{Bejan+08} and its extended version ECB+~\cite{Cybulska+14}, the data released at the relatively recent evaluation campaign Knowledge Base Population (KBP) track at Text Analysis Conference (TAC)~\cite{Getman+18}, OntoNotes~\cite{Pradhan+2007}, and Rich ERE~\cite{Song+15} are the main datasets that contain explicit annotations for event coreference. Although, many event types are covered in these datasets, the coverage is generic in terms of event types and the focus is on linguistic aspects of event manifestations. The analysis of the nuances and context dependent characteristics such as the prevalence in a random sample of news of protest events is not in the scope of these datasets 

Majority of the event coreference corpora consists of documents in English. A few of the available datasets are mainly in English and incorporate data in other languages such as Chinese~\cite{Doddington+2004,Getman+18}, Catalan~\cite{Recasens+12}, and Spanish~\cite{Huang+16,Getman+18} as well.~\footnote{A detailed survey of the event coreference datasets is reported by~\citet{Lu+18}.} 

The task event coreference resolution has not been in the scope of the studies of the social scientists that work on automated event data collection. The few protest event corpora proposed by \citet{Sonmez+16}, \citet{Makarov+16}, and \citet{Hurriyetoglu+21} do not include event coreference information. The sentence level coreference information was the focus of the shared tasks Event Sentence Coreference Identification (ESCI) in English \cite{Hurriyetoglu+20b} and in English, Portuguese and Spanish \cite{hurriyetoglu-etal-2021-multilingual}. Although it is about protest events, work by \citet{Huang+16} focuses only on temporal status of the events, which can be past, on-going, and future.

We propose the first multilingual corpus for protest event coreference resolution. The other unique features of the corpus are being based on random sampling and active learning and containing news articles that report a single event using a single trigger as well. These features enable us to understand manifestation of events in a representative text collection and improve the methodology for protest event information collection by highlighting the importance of the event coreference in real world event information collection studies. Last but not least, availability of this corpus will contribute to the development and evaluation of event information collection systems usable in real scenarios \cite{giorgi-etal-2021-discovering}. 

\section{Protest Event Definition}
\label{sec:protest-definition}

We define a protest as “a collective public action by a non-governmental actor who expresses criticism or dissent and articulates a societal or political demand”~\cite{rucht1999acts} (p. 68), and instances or episodes of social conflict, which are based on grievances or aspirations to change the social and political order. These events are referred as the ``repertoire of contention'' in the scope of contentious politics~\cite{Giugni98,Tarrow94} and most similar to~\citet{Sonmez+16} and~\citet{Makarov+16} in the context of our study. We operationalize this definition as contentious politics event (CPE) and use the term protest interchangeably with it throughout this paper.

CPEs cover any politically motivated collective action which falls outside the official mechanisms of political participation associated with formal government institutions of the country in which the said action takes place. This broad event definition is developed and fleshed out on two levels. First we identify three abstract categories of collective action, namely, political mobilizations, social protests, and group confrontations, in order to define the broad range of events that we focus on. Next, five specific categories of CPEs are identified as concrete manifestations of these three modes of collective action. Demonstrations (rallies, marches, sit-ins, slogan shouting, gatherings etc.), industrial actions (strikes, slow-downs, picket lines, gheraos etc.), group clashes (fights, clashes, lynching etc.), armed militancy (attacks, bombings, assassinations etc.) and electoral politics events (election rallies) are the concrete types of events our event ontology encompasses. 

We define criteria to which the news stories that report protest events must conform in order to be classified as relevant. The criteria are the necessity of civilian actors, and the existence of concrete or implied time and place information which ascertains that the event(s) has definitely taken place. Only reports that mention events that took place in the past, or are taking place at the time of writing are labeled as protest news articles. The references to the future (i.e. planned, threatened, announced or expected) events are not labeled as protest, with the exception of threats of or attempts at violent actions.\footnote{Although planned events and protest threats could have a role in our analysis~\cite{Huang+16}, they are neither relevant in the protest reporting context nor their prevalence, which is below 0.5\% of a random sample according to our observations, allow their automated analysis.} The comparison of our definition with ACE event ontology~\cite{Doddington+2004} is provided in Appendix~\ref{appendix:protest-vs-ACE}.



Events are annotated for their semantic types as well. The event types are 
\begin{description}
 \item[Demonstration] A demonstration is a form of political action in which a demand or grievance is raised outside the given institutionalised forms of political participation in a country. 
 \item[Industrial action] Industrial actions are events that take place within workplaces or involve the production process in the protest. 
 \item[Group clashes] Group clashes are confrontations that stems from politicized conflicts (e.g. identity or economic interest based or ideological conflicts) between social groups
 \item[Armed militacy] Politically motivated violent actions that fall within our event definition are included in this category. 
 \item[Electoral politics] These events are rallies, marches or any similar mass mobilizations that are organized within the scope of election campaigns of political parties or leaders.
 \item[Other] Any CPE which does not fit in one of the categories listed above is marked with this tag.
 
\end{description}



\section{Event Separation for Coreference Annotation}
\label{sec:event-sep}

If an event is referred with multiple words in a sentence, these mentions are marked as coreferent. This is the case in \textit{Ex1} and \textit{Ex2} in Table~\ref{tab:ec-examples}. Coreferent event mentions may occur across sentences as well, e.g. \textit{Ex3}. The news articles may report more than one event and pieces of information about one event might not be applicable to the other event. In this case, we need to distinguish different events within the article and link the arguments to the correct event mentions. This is referred to as event separation and is subject to a number of rules to ensure coherence in annotation. Note that in separating events we need to think of the news text rather than the actual reality that the text recounts. That is to say, we are more interested in the separate event references in the text than whether the said events are actually separate from each other in real life. As will be clearer in the examples demonstrated in Table ~\ref{tab:ec-examples}, sometimes it is not possible to know or show for certain whether separate event references correspond to separate real life events. For instance, there are two separate events in \textit{Ex4}. BJP workers’ demonstration is the first event and the attack at the train station is the second event (in the order in which they appear in the document). 

\begin{table*}[!th]
\rowcolors{2}{}{black!10!}
\centering
\resizebox{\textwidth}{!}{
\begin{tabular}{p{.99\textwidth}}
\toprule 
\midrule
\textbf{Ex1:} The students organized a \textbf{(e\textsubscript{1}: protest)} by \textbf{(e\textsubscript{1}: marching)} against the payment seat decision. \\
\textbf{Ex2:} Commenting on the \textbf{(e\textsubscript{1}: strike)} which was flagged off on Monday, the union secretary stated “\textbf{(e\textsubscript{1}: it)} will continue as long as our demands are not met. \\
\textbf{Ex3:} CPI(M) stages \textbf{(e\textsubscript{1}: protest)} rally in Bhavnagar. The Bhavnagar unit of communist party of India CPI(m) on Friday staged a \textbf{(e\textsubscript{1}: demonstration)} opposite the local post office here. \\ 
\textbf{Ex4:} At noon, BJP workers \textbf{(e\textsubscript{1}: gathered)} in the square and shouted slogans, condemning the failure of the Union Government in delivering justice to the victims of last year’s terror \textbf{(e\textsubscript{2}: attack)} at the train station where armed militants killed 25 people. \\
\textbf{Ex5:} Karnataka State Government Employees Association organized \textbf{(e\textsubscript{1,2}:demonstrations)} in Bangalore and Mysore yesterday, urging the government not to go ahead with the new retirement scheme.\\
\midrule
\bottomrule
\end{tabular}}
\caption{Event coreference examples i) \textit{Ex1}, \textit{Ex2}, and \textit{Ex3} contain event triggers that express the same event, ii) The triggers in \textit{Ex4} are about separate events, and iii) The trigger in \textit{Ex5} denotes events that take place in Bangalore and Mysore.
}\label{tab:ec-examples}
\end{table*}

The separation of event references is based on difference in at least one of the following:
\begin{description}
\item [Time] Events that occur at different times are separated from each other. The time difference necessary for separation is 24 hours. Events that continue throughout the same day are not separated even if they are reported to occur at different times of the day.
\item [Location] Events which are reported to take place in different locations are separated as different events. Locations can be event places or facilities. An event that has started at some place and continued at another, e.g. a march that started at a location and proceeded at somewhere else, is not separated. However, if an event is happening simultaneously at multiple locations or at multiple locations at different times but not in continuum are separated such that every location reference count as a separate event. Demonstrations in Bangalore and Mysore are annotated as belonging to separate events in \textit{Ex5}, although they share the event trigger \textit{demonstration}.
\item[Participant or organizer] Events which are carried out by different participants or organizers with separate goals and motivations are separated. This separation takes place even in cases where different protests occur at the same time and location. The separation is based on event motivations or goals but since motivation info is not something that we annotate and might at times be elusive, we distinguish events based on participants and organizers. The most frequent cases which exemplify this situation are that of counter-protests where two groups of participants or organizers demonstrating against each other and/or with conflicting agendas. Note that in cases where there are multiple types of participants and/or organizers that protest together, the event will not be separated.
\item [Semantic event category] Events which occur at the same time, place and facility, and organized and participated by the same participants but have a different semantic category are separated as different events. In other words, as a result of this, the triggers of each event in a document that is separated by its respective event number will have only one semantic category tag. Although this case is rare, it can be encountered when rallies, marches or other types of demonstrations occur during industrial strikes.
\end{description}

Event information can be spread over a document and occur in a sentence that does not contain the respective event mention. This event information is not annotated. This is to say only event information that co-occur with the related event mention in a sentence is annotated. Moreover, there might be event triggers (types or mentions) that are plural such as \textit{Ex5}, i.e. refer to more than one event that are separated. We have a unique procedure for separating these events. In a nutshell, if an article contains a plural event reference, such as ``protests'' which refers to e.g. two different events, each of which are reported in the article, that article will have three separate event numbers. This is because, the event reference "protests" is counted as an event reference on its own.\footnote{The reason for this is that, a plural event mention might have different arguments from the singular events that it designates. For instance, in the sentence "The plaza was the scene of protests for the last two weeks" the reference "protests" has the time argument "last two weeks". The references to events that make up these "protests" will have their corresponding and distinct time arguments elsewhere in the article, as in, "last week", and "the week before last week".}

\section{Methodology for Corpus Creation}
\label{sec:methodology-corpus-create}
A corpus that has the capacity to support creation of automated systems for event information collection in the wild must be representative of the event type occurrence in real life~\cite{halterman-etal-2021-corpus,Yoruk+2021}. Therefore, our corpus is based on a randomly sampled news articles from online archives of local news sources from India, China, South Africa, Argentine, and Brazil. English data was collected from \textit{The Hindu}, \textit{South China Morning Post (SCMP)}, \textit{New Indian Express}, \textit{Indian Express}, \textit{Guardian}, and \textit{African News Agency} journals. The news articles in Portuguese were retrieved from \textit{Folha} and \textit{Estadao}. Finally, the Spanish documents were gathered from \textit{Clarin}, \textit{Pagina12}, and \textit{La Nacion}. The news archives mainly cover the period between 2000 and 2019. Although the majority of the documents are from random samples, we facilitated a high recall active sampling to extend the random samples in cases they do not contain sufficient number of positive samples for modelling protest events.

The annotation starts with labelling the articles as containing a protest event or not. Next, the same procedure is applied on the sentences of the documents that are ensured to have protest information by applying adjudication, spotcheck, and error correction. Both at the document and sentence levels, at least one event trigger must occur in the instance to qualify for the positive label. The positively labeled sentences are annotated at token level for event triggers, arguments such as time, place, and event actors, and semantic category of these event triggers. Finally, the event triggers are connected to each other in case they are about the same event. Document and sentence level labelling is applied on an online tool we have developed in-house. The event sentence grouping and token level annotations are performed utilizing FLAT.\footnote{\url{https://github.com/proycon/flat}, accessed on October 10, 2021.}. Annotators always see complete documents and any annotations that are agreed upon from previous level(s).

We pay particular attention to the quality of the annotations. Detailed annotation manuals were prepared and updated as they are tested against the data. Each annotation on an instance at any level is performed by two graduate students who are studying social or political science and trained on the annotation methodology. Moreover, they were trained about the socio-political context of the country the news articles to be annotated. Therefore, if a news article reports on an event that had not occurred in the target country, this article is only labelled at document and sentence levels. But it is not included in the event coreference dataset. The English text from India, China, and South Africa was annotated by a team of annotators whose native language is Turkish and living in Turkey. The annotations on Spanish and Portuguese text from Argentine and Brazil respectively was prepared by a team of annotators whose native language is Portuguese and live in Brazil. 

Disagreements between annotators are adjudicated by the annotation supervisor, who is a political scientist and responsible for maintaining annotation manuals for each annotation task, such as document labelling, sentence labelling, and token level event annotation. The annotation supervisor performs a spotcheck to around 10\% of the agreements. Finally, for each task semi-automated quality checks were performed by using the adjudicated data for both training and testing a machine learning model. The disagreements between the predictions and annotations were analyzed by the annotation supervisor. The quality enhancement efforts has enabled us to update around 10\% of all of the annotations.

\section{Corpus Characteristics}
\label{sec:corpus-character}



The corpus consists of documents in English (EN), Portuguese (PR), and Spanish (SE), which are represented with 896, 97, and 106 documents respectively. The inter-annotator agreement (IAA) was measured using Krippendorf's alpha~\cite{Krippendorff+16} for the document, sentence, and token level annotations. Table~\ref{table:iaa-scores} provides the average IAA scores in the rows \textit{Document}, \textit{Sentence}, and \textit{Token} for each language. The columns \textit{Time}, \textit{Trigger}, \textit{Place}, \textit{Facility}, \textit{Participant}, \textit{Organizer}, and \textit{Target} break down the average \textit{Token} scores. The IAA for event coreference annotation was measured by comparing labels of the annotators with the adjudicated annotations using scorch - a Python implementation of CoNLL-2012 average score for the test data~\citep{Pradhan+14}.~\footnote{\url{ https://github.com/LoicGrobol/scorch}, accessed on October 28, 2021.} The scores for EN, PR, and ES are 88.58, 89.72, and 68.64.

The IAA score for some of the token level annotations are relatively low. This can be speculated to be caused by the native language of the annotators, which is provided in the \textit{Native} row in Table \ref{table:iaa-scores}. The quality assurance steps that are 100\% double annotation, adjudication of all disagreements, spotcheck of the 10\% of the annotations agreed on, and semi-automated annotation error correction ensure the low IAA scores not to affect the utilization of the corpus. 


\begin{table}
\centering
\resizebox{.47\textwidth}{!}{\begin{tabular}{lccccccc}
\toprule
 & English & Portuguese & Spanish \\
\midrule
Document & .75 & .82 & .83 \\
Sentence & .65 & .72 & .79 \\
Token & .39 & .48 & .39 \\
\midrule
Time         & .59 & .52 & .53 \\
Trigger       & .38 & .44 & .45 \\
Place        & .41 & .47 & .49 \\
Facility         & .34 & .42 & .32 \\
Participant       & .36 & .51 & .39 \\
Organizer        & .45 & .67 & .26 \\
Target        & .25 & .41 & .25 \\
\midrule
Native & Turkish & Portuguese & Portuguese \\
\bottomrule
\end{tabular}}
\caption{The inter-annotator agreement for document, sentence, and token levels in terms of Krippendorff's alpha. Token level scores are provided for the trigger and its arguments as well. Finally, the row \textit{Native} provides the native language of the annotation teams.}
\label{table:iaa-scores}
\end{table}

Table~\ref{table:ec-lang-stats} demonstrates the number of documents, sentences, and event mentions in the rows \textit{\#docs}, \textit{\#sents}, and \textit{\#events} for English (EN), Portuguese (PR), and Spanish (SE) respectively. Moreover, the Table provides information on the amount of event information that could be identified precisely under the assumptions \begin{inparaenum}[1)]
\item a document contain information about a single event,
\item a sentence contain information about a single event, and
\item information about an event is reported in a single sentence.
\end{inparaenum} The first assumption could capture the information presented in \textit{\#docs1e} which shows it holds for 532 (59.38\%), 60 (61.86\%), and 68 (64.15\%) documents. The average number of events in a news articles that reports a protest event is two. The second allow 3,255, 320, and 404 out of 3,559, 352, and 449 sentences to be processed based on this assumption respectively. Around 10\% of the sentences contain mentions of multiple separate events, which is around 15\% of the total event information. The third is valid only for 763 (46\%), 86 (47.77\%), and 82 (44.80\%) of the events. Although the documents that contain information about a single event are more than the ones that contain event information about multiple events, more than half of the event information occur in documents that contain information about multiple events. 

Last but not least, the event mentions that refer to more than one event is around 9\% across all languages.

\begin{table}
\centering
\resizebox{.40\textwidth}{!}{\begin{tabular}{lccc}
\toprule
 & EN & PR & SE \\
\midrule
\#docs         & 896 & 97 & 105 \\
\#docs1e       & 532 & 60 & 68 \\
\midrule
\#sents        & 13,584 & 1,397 & 2,669 \\
\#esents       & 3,559 & 352 & 449 \\
\#sents1e      & 3,255 & 320 & 404 \\
\midrule
\#events       & 1,651 & 180 & 183 \\
\#events1sent & 763 & 86 & 82 \\
\bottomrule
\end{tabular}}
\caption{The number of documents (\#docs), sentences (\#sents), and events (\#events) in English (EN), Portuguese (PR), and Spanish (SE). Documents and sentences that contain information about one event (\#docs1e and \#sents1e) and events mentioned only in one sentence (\#events1sent) show the prevalence of event coreference.}
\label{table:ec-lang-stats}
\end{table}

We have created train, validation, and test splits that has the ratio .70, .15, and .15 respectively in order to facilitate experimentation, benchmarking, and reproduciability. The splits are presented in Table~\ref{table:data-splits}. The ratio, which is provided in the row \textit{Positive ratio}, of the documents that contain events is more or less the same across splits in a language.

\begin{table}
\centering
\resizebox{.30\textwidth}{!}{\begin{tabular}{lccc}
\toprule
 & EN & PR & SE \\
\midrule
\#train         & 628 & 67 & 74 \\
\#validation    & 134 & 15 & 16 \\
\#test          & 134 & 15 & 16 \\
\midrule
Positive ratio  & .58 & .59 & .53 \\
\bottomrule
\end{tabular}}
\caption{The number of documents in the train, validation, and test splits for English (EN), Portuguese (PR), and Spanish (ES). The ratio of the documents that contain events is provided in the row \textit{Positive ratio}.}
\label{table:data-splits}
\end{table}

\section{Event Coreference Resolution Methodology}
\label{sec:event-coref-resolution}

\begin{figure*}
\begin{center}
\includegraphics[scale=0.45,trim=0 60 0 55, clip]{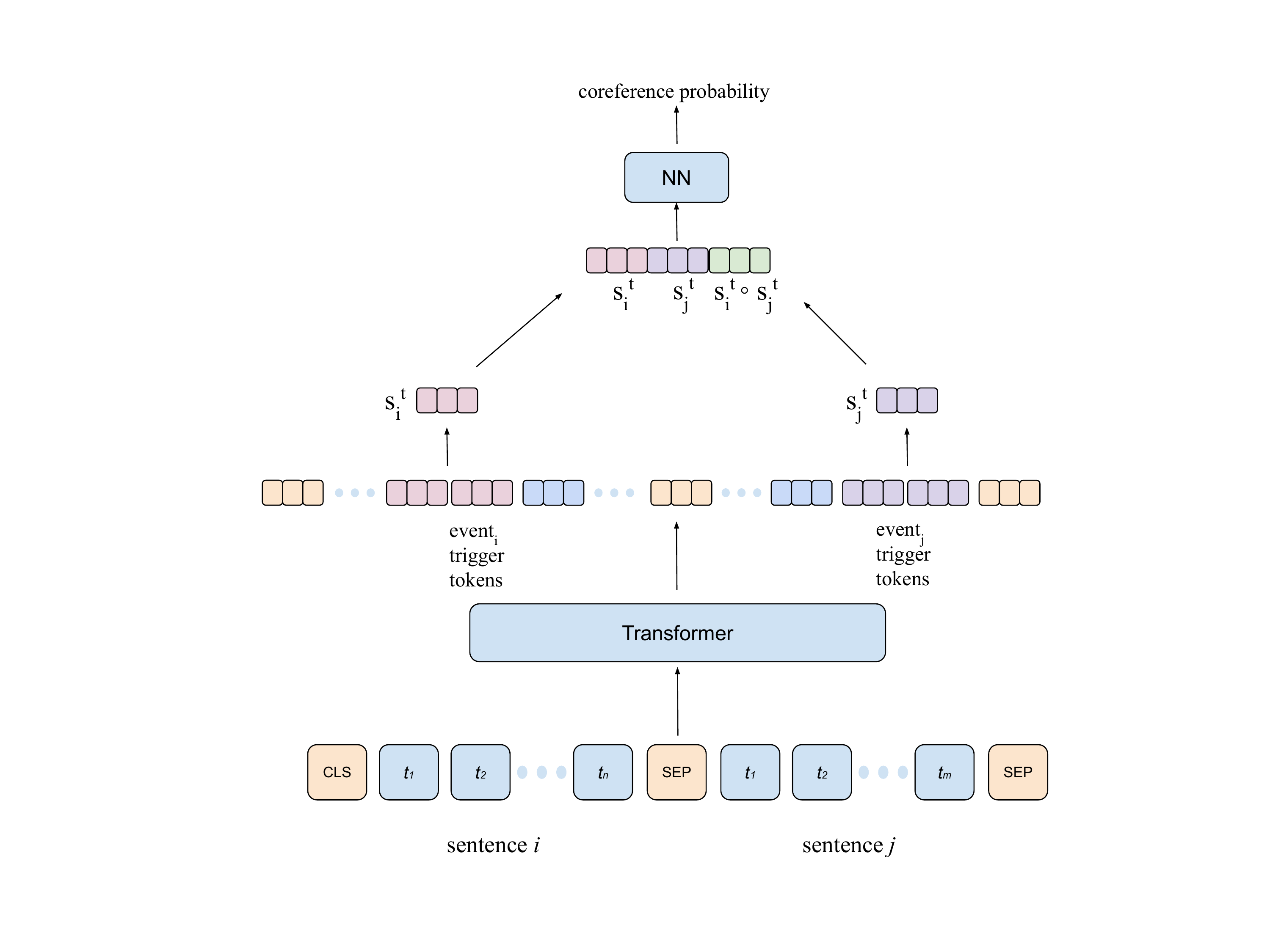} 
\caption{The architecture that was proposed by~\citet{Yu2020PairedRL}. The sentence pairs are fed to the transformer model to get token embeddings. To obtain the final trigger vector for a given event mention, the point-wise average of token representations, which are part of the trigger span, is calculated. These tokens might come from multiple words or as subtokens of a single word. Lastly, the trigger vectors are concatenated with their point-wise multiplication to compose the final representation of trigger pairs in sentences \textit{i} and \textit{j}. The final representation is fed into a two-layer multi layer perceptron (MLP) that yields the probability of being coreferent for a given trigger pair.} 
\label{fig:architecture}
\end{center}
\end{figure*}

We evaluated performance of a state-of-the-art monolingual and multilingual transformer models in an architecture proposed by~\citet{Yu2020PairedRL}, which is illustrated in Figure~\ref{fig:architecture}, on the corpus. Moreover, we have calculated a dummy baseline score on the validation and test data. The baseline predicts all events as being in the same cluster in a document, i.e., maximum cluster prediction (MaxC). This baseline is the reflection of assuming a document contains information about a single event.

In addition to use the standard threshold, which is .50 for predicting coreference relation, we optimized it by evaluating all values starting from .01 until .99 by increasing the threshold by .01 as a threshold on the validation set for each language. 

Neither the models nor the baseline fully utilize the event information that occurs in event mentions that refer to multiple events and sentences that contain event mentions about more than one event. The event label that occurs more than other event labels assigned to an event mention is the final label of the event mention. In case the occurrence frequency of the assigned event labels are the same, the one that occurs first is used. 

The sentences that contain more than 512 tokens are ignored if all event mentions are not in the first 512 tokens. This was the case in only nine sentence pairs in Spanish training data.~\footnote{The models we have created can be found on \url{https://www.dropbox.com/sh/7j2j3f06kbn5ziv/AACVvvoFe5HH52PSKWTLph2Oa?dl=0}}

\section{Results}
\label{sec:results}

The transformer models utilized are SpanBERT~\cite{Lu-Ng-SpanBased-2021}\footnote{\url{https://huggingface.co/SpanBERT/spanbert-base-cased}, accessed on November 15, 2021.}, RoBERTa~\cite{Liu-roberta-2019}\footnote{\url{https://huggingface.co/roberta-base}, accessed on November 15, 2021.}, and mBERT~\cite{Devlin+19}\footnote{\url{https://huggingface.co/bert-base-multilingual-uncased}, accessed on November 15, 2021.}. The training data is set as English and validation and test data is the respective subsets in each language.\footnote{Although they are not used to train the models for Portuguese and Spanish, the splits are provided for all languages as we believe these splits are critical for benchmarking purposes.}

Table~\ref{tab:results} demonstrates the performance of MaxC, SpanBERT, and RoBERTa on the validation and test sets. The multilingual modeling is achieved using mBERT. All scores are generated using a single random seed, which is 44, and measured utilizing scorch for the scores in terms of F1, MUC, B\textsuperscript{3}, CEAF\textsubscript{e}, Blanc, and CoNLL 2012. The CoNLL 2012 score is used for comparing the systems as it is the average of MUC, B\textsuperscript{3}, and CEAF\textsubscript{e} as each of the three metrics represents a different aspects of the performance~\cite{Pradhan+12}

\begin{table*}[!htbp]
\centering
\resizebox{.97\textwidth}{!}{\begin{tabular}{|l|c|cccccc|cccccc|}
\hline
   & thres & \multicolumn{6}{c}{Validation} & \multicolumn{6}{c}{Test} \\
   & & F1 & MUC & B\textsuperscript{3} & CEAF\textsubscript{e} & Blanc & CoNLL & F1 & MUC & B\textsuperscript{3} & CEAF\textsubscript{e} & Blanc & CoNLL \\
    \hline
    MaxC\textsubscript{EN} & - & 73.48 & 90.64 & 82.64 & 60.57 & 86.62 & 77.95 & 72.80 & 91.75 & 82.76 & 62.79 & 86.41 & 79.10 \\
    \multirow{2}{*}{SpanBERT} & .50 & 79.94 & 89.86 & 83.75 & 68.52 & 86.13 & 80.71 & 80.06 & 91.11 & 84.20 & 71.42 & 85.84 & 82.24 \\
    & .53 & 79.71 & 90.00 & 83.93 & 69.43 & 86.07 & 81.12 & 79.95 & 90.90 & 84.13 & 71.49 & 85.96 & 82.18 \\
    \multirow{2}{*}{RoBERTa} & .50 & 80.83 & 91.07 & 84.12 & 65.99 & 87.17 & 80.39 & 81.33 & 93.04 & 85.21 & 70.20 & 88.13 & \textbf{82.82} \\
    & .54 & 81.00 & 91.28 & 84.15 & 66.44 & 86.98 & 80.62 & 81.52 & 92.94 & 85.03 & 69.87 & 87.99 & 82.61 \\
    \multirow{2}{*}{mBERT\textsubscript{EN,EN}} & .50 & 77.73 & 90.51 & 83.27 & 65.19 & 85.92 & 79.66 & 80.38 & 92.14 & 84.63 & 70.05 & 86.91 & 82.27 \\
    & .87 & 76.32 & 89.89 & 82.95 & 66.62 & 85.06 & 79.82 & 79.60 & 91.44 & 84.74 & 71.44 & 85.85 & 82.54 \\
    \hline\hline
    MaxC\textsubscript{PR} & - & 74.27 & 93.80 & 85.38 & 72.57 & 86.17 & 83.92 & 72.07 & 89.07 & 79.24 & 58.21 & 84.95 & 75.51 \\
    \multirow{2}{*}{mBERT\textsubscript{EN,PR}} & .50 & 78.69 & 94.64 & 86.35 & 75.88 & 86.36 & 85.62 & 77.23 & 92.03 & 84.59 & 68.66 & 87.04 & 81.76 \\
    & .56 & 78.93 & 94.64 & 86.35 & 75.88 & 86.36 & 85.62 & 77.23 & 92.03 & 84.59 & 68.66 & 87.04 & \textbf{81.76} \\
    \hline\hline
    MaxC\textsubscript{ES} & - & 68.92 & 89.41 & 74.93 & 45.37 & 82.63 & 69.89 & 66.86 & 91.78 & 79.39 & 58.85 & 81.95 & \textbf{76.67} \\
    \multirow{2}{*}{mBERT\textsubscript{EN,ES}} & .50 & 73.55 & 90.47 & 77.41 & 50.91 & 83.72 & 72.93 & 67.38 & 90.27 & 77.50 & 54.81 & 79.78 & 74.20 \\
    & .97 & 73.86 & 89.99 & 78.74 & 52.23 & 80.78 & 73.65 & 64.44 & 88.73 & 76.32 & 54.13 & 78.63 & 73.06 \\
    \hline
\end{tabular}}
\caption{Baseline and transformer model performances for event coreference resolution on our corpus. \textit{MaxC} is the baseline calculated by assuming all event mentions in a document refer to the same event. SpanBERT and RoBERTa are trained and tested using respective splits of the English data. mBERT is trained using the English training data and validated and tested on the target language, which is Portuguese for mBERT\textsubscript{EN,PR} and Spanish for mBERT\textsubscript{EN,ES}. The \textit{thres} column is the probability threshold for determining whether two event mentions are coreferent.}
\label{tab:results}
\end{table*}

Although, RoBERTa has obtained the best CoNLL 2012 score, which is 82.82, on the English test set, the results of SpanBERT are comparable. The threshold optimization does not help any of these two models. The performance of mBERT\textsubscript{EN,EN} that is trained and validated on the respective splits of the English data is slightly higher than SpanBERT and RoBERTa. The mBERT models that are trained on English data and validated and tested on respective splits of Portuguese and Spanish data is reported in the rows mBERT\textsubscript{EN,PR} and mBERT\textsubscript{EN,ES} respectively. mBERT\textsubscript{EN,PR} outperforms the baseline by obtaining 81.76 CoNLL 2012 score. However, threshold optimization on validation set does not improve performance on test data. Finally, the performance of mBERT\textsubscript{EN,ES} remain below the baseline even after threshold optimization.



\section{Conclusion}
\label{sec:conclusion}

We have explored the prevalence of event coreference in a random sample of news articles collected from multiple sources, languages, and countries. We have found that the news articles contain two events in average and state-of-the-art transformer models can improve determination of separate events in most of the evaluation scenarios.

We aim at tackling multilingual event coreference resolution by first testing and improving the work reported by \citet{phung-etal-2021-learning} \citet{awasthy-etal-2021-ibm}, and \citet{tan-etal-2021-nus} on our dataset.

\section*{Acknowledgments}
The authors from Koc University are funded by the European Research Council (ERC) Starting Grant 714868 awarded to Dr. Erdem Y\"{o}r\"{u}k for his project Emerging Welfare. 

\bibliography{emwwp2,anthology,custom}
\bibliographystyle{acl_natbib}

\appendix

\section{Comparison of our Protest Event Definition with ACE Event Ontology}
\label{appendix:protest-vs-ACE}

The ATTACK and DEMONSTRATE categories in the CONFLICT heading of the ACE English Annotation Guidelines for Events coding manual \cite{Doddington+2004}\footnote{\url{https://www.ldc.upenn.edu/sites/www.ldc.upenn.edu/files/english-events-guidelines-v5.4.3.pdf}, accessed on October 10, 2021}, have commonalities with our event ontology, however, they are not applicable in the latter setting due to fundamental differences between how events are defined in the two annotation schemes. ACE annotation principles define events as any ``specific occurrence involving participants. An event is something that happens'' (p.5). This abstracts the actors from the definition, making event type and sub-type definitions neutral in terms of actors. Namely, ACE event type labels are employed based solely on the nature of the occurrences -``acts'' in relevant types- regardless of the nature of participants. On the other hand, our event ontology focuses on CPEs, which, by their nature, involve a particular type of actor from the outset, namely, civilian, that is non-state actors. In this respect, the ATTACK event type, which is defined as any ``violent physical act causing harm or damage'' (p.33) in ACE event coding rules, is not applicable in CPE coding as it includes state actions, such as international wars and military actions against non-state actors. In other words, despite many event examples of the ATTACK type enumerated in ACE manual, such as ``attack'', ``clash'', ``bomb'', ``explode'', overlap with our event definition, they will be excluded from the latter when their authors are state actors due to their different, non-contentious politics nature. 

The second similar event type category in ACE event annotation guidelines is the DEMONSTRATE category. It is defined as including events that occur ``whenever a large number of people come together in a public area to protest or demand some sort of official action'' (p.34). This definition is better aligned with the CPE ontology we define due to the fact that it designates actions of social and/or political actors that are non-state. However, this definition, in itself, is too restrictive to be applicable in terms of a broad understanding of contentious politics for two reasons. First, as it seems to limit the scope of this event type to spontaneous (that is unorganized) gatherings of people, it excludes certain actions of political and/or grassroots organizations such as political parties and NGOs. Protest actions of such organizations sometimes do not involve mass participation despite aiming at challenging authorities, raising their political agendas or issuing certain demands. Putting up posters, distributing brochures, holding press declarations in public spaces are examples of such protest events. Secondly, the requirement of mass participation in a public area leaves many protest actions such as on-line mass petitions and boycotts, which are not necessarily tied to specific locations where people actually gather, and actions of individuals or small groups such as hunger strikes and self-immolation. Due to the fundamental incompatibilities detailed above, we opted to develop a specific event ontology and annotation guidelines\footnote{The detailed guidelines will be provided either as supplementary material or upon acceptance of the paper.} that are different from event definitions in ACE guidelines.

\section{Reproduciability notes}
\label{appendix:reproduciability}

The following libraries were utilized to conduct the experiments: python == 3.8.10, torch == 1.9.0, pytorch\_lightning == 1.3.8, and transformers == 4.8.2\\
\\The following hyperparameters are optimized:

\begin{description}
\item[Threshold] The probability of being coreferent for two event mentions are tested from .01 to .99 by incrementally increasing the threshold by .01.
\end{description}
\begin{description}
\item[Learning Rate] Each model was trained using the learning rate of 5-e6 which was searched in \{1-e5, 5-e5, 1-e6, 5-e6, 1-e7\}.
\end{description}
\begin{description}
\item[AdamW Eps] We used AdamW optimizer for our models. Eps value for our optimizer was selected as 1-e6 which was searched in \{1-e6, 1-e7, 1-e8\}
\end{description}
\begin{description}
\item[Hidden Unit] Each model used used identical classifier heads which was a two-layer MLP. 128 was the selected hidden unit which was searched in {32, 64, 128, 256}.
\end{description}

We have used fixed parameters for each model.\\
\\The number of epochs needed for each model to be trained is 2 to get shared baseline results. The average run-time for an epoch is 10 minutes.\\
\\All experiments were performed on the same machine with 10 Intel i9-10900X CPUs, and 2 NVIDIA RTX 2080 (8 GB) GPUs. We did not perform distributed training among the GPUs. Full memory of a single GPU was enough to perform each experiment.
\end{document}